\documentclass[11pt]{article}
\usepackage[top=0.7in, bottom=0.7in, left=1.2in, right=1.2in]{geometry}

\usepackage{amsmath, amssymb, graphicx}
\usepackage{hyperref}
\usepackage{graphicx}
\usepackage{media9}
\usepackage{float}
\title{Advancing Employee Behavior Analysis through Synthetic Data: Leveraging ABMs, GANs, and Statistical Models for Enhanced Organizational Efficiency}
\author{Rakshitha Jayashankar, Mahesh Balan}
\date{}

\begin{document}

\maketitle

\begin{abstract}
 Success in today's data-driven corporate climate requires a deep understanding of employee behavior. Companies aim to improve employee satisfaction, boost output, and optimize workflow. This research study delves into creating synthetic data, a powerful tool that allows us to comprehensively understand employee performance, flexibility, cooperation, and team dynamics. Synthetic data provides a detailed and accurate picture of employee activities while protecting individual privacy thanks to cutting-edge methods like agent-based models (ABMs), Generative Adversarial Networks (GANs), and statistical models. Through the creation of multiple situations, this method offers insightful viewpoints regarding increasing teamwork, improving adaptability, and accelerating overall productivity. We examine how synthetic data has evolved from a specialized field to an essential resource for researching employee behavior and enhancing management efficiency.\\
\textbf{Keywords}: Agent-Based Model, Generative Adversarial Network, workflow optimization, organizational success
\end{abstract}

\section{Introduction}
Employee behavior in the workplace is not just a factor but a cornerstone that significantly influences the organization's culture, productivity, and success. In people analytics, employees' active and energetic behaviors are helpful and crucial for organizations to achieve their goals[1]. Synthetic data is artificial data produced to mimic accurate data using computational methods and simulations for various uses like testing, model training, and research. To achieve organizational goals and objectives, employees are crucial in utilizing this and other resources, such as technology, finance, information, and other managerial tools[2]. The complexity of synthetic data generation can range from simple mathematical equations to fully simulated virtual environments[3]. The use of artificially created data in the field of employee-organizational structure helps maintain sensitive information in the analysis without exposing employee details, thus safeguarding employees' privacy and complying with data protection regulations. This artificially created data can be generated from an actual data set, or a completely new dataset can be generated if the actual data is unavailable. The newly generated data is nearly identical to the original data. In this research paper, we used this method to obtain employee information through some open-source platforms and generated employee-specific data using methodologies.

\section{Advantages of Synthetic Data in the Field of People Analytics}
\subsection{Privacy and Realism}
Mimicking the real-world data scenario, generating artificial data, and replicating the data with the same statistical properties, distributions, and relationships as actual data. Direct identifiers like names, employee IDs, or contact information are removed or replaced with random identifiers. Grouping similar characteristics reduces the granularity of the data, making it harder to trace back to specific individuals.
\subsection{Scalability and Bias-free Generation}
Synthetic data generation can create large, unbiased datasets, even in data-sparse areas, which is a significant advantage in people analytics. It allows for building large, representative datasets that mirror potential employee behaviors, filling in gaps where accurate data is sparse.
\subsection{Cost efficiency and versatility}
Generating synthetic data is a cost-effective solution significantly cheaper than collecting large amounts of real-world data. This cost efficiency brings financial benefits to the organization's data-driven projects. The versatility lies in the ease with which it can be modified or regenerated to meet specific needs.

\section{Methodology}
\subsection{Statistical Methods}
\subsubsection{Multivariate Method}
A statistical method examining multiple variables and their interactions simultaneously captures their interdependencies. This method helps generate data that maintains the characteristics of the original dataset.
It models a joint distribution of multiple normally distributed variables. This captures the correlations between multiple variables, ensuring a reliable data generation process. The data follows a normal distribution, making it straightforward and efficient. 
When applied to employee behavior, where factors such as performance, team engagement, collaboration, and flexibility are involved, the multivariate method allows us to generate synthetic data that mimics the real-world relationships between these factors.
\[
\text{Cov}(X, Y) = \mathbb{E}\left[(X - \mathbb{E}[X])(Y - \mathbb{E}[Y])\right]
\]

\[
\text{Corr}(X, Y) = \frac{\text{Cov}(X, Y)}{\sigma_X \sigma_Y}
\]

\text{where } \(\sigma_X\) \text{ and } \(\sigma_Y\) \text{ are the standard deviations of } \(X\) \text{ and } \(Y\) \text{ respectively.}

\begin{figure}[H] 
  \centering
  \includegraphics[width=0.85\textwidth]{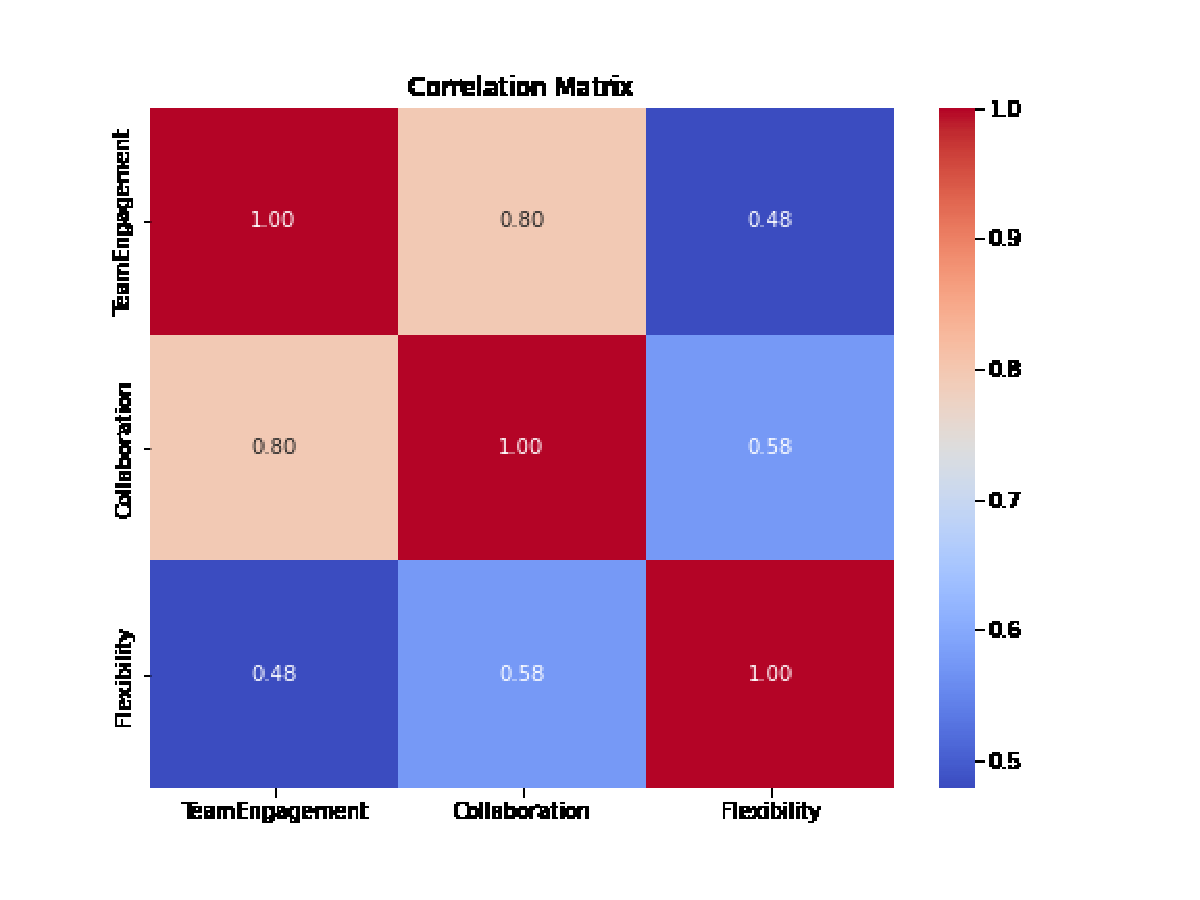}
  \caption{Heatmap for Multivariate Method}
\end{figure}
A Strong correlation (0.8) between  Team Engagement and collaboration indicates that collaboration naturally improves in work environments where teams are more engaged.
The moderate correlations suggest that while Flexibility is positively associated with Team Engagement and Collaboration, the relationship is not as strong. This indicates that flexibility might be influenced by other factors not represented in this matrix or has a less direct relationship with engagement and collaboration.

\subsubsection{Bootstrapping Method}
Introduced by Bradley Efron in 1979, the bootstrapping method is a non-parametric approach to statistical interference. This method involves creating multiple “bootstrap samples” from the original dataset by resampling with replacement. By resampling the original data with replacement and calculating the statistic of interest across many bootstrap samples, we can confidently estimate the variability of the statistic. This ability to accurately estimate variability is a key strength of the method, allowing you to construct confidence intervals, test hypotheses, or assess model stability with a high degree of reliability.
Bootstrapping involves resampling the performance data with replacement to create synthetic data. Noise is added to each variable to make the data more realistic.
\[
Z_{ji} = y_{ji} + e_{ji}, \quad e_{ji} \sim \mathcal{N}(0, \sigma^2)
\]
\begin{figure}[H] 
  \centering
  \includegraphics[width=0.85\textwidth]{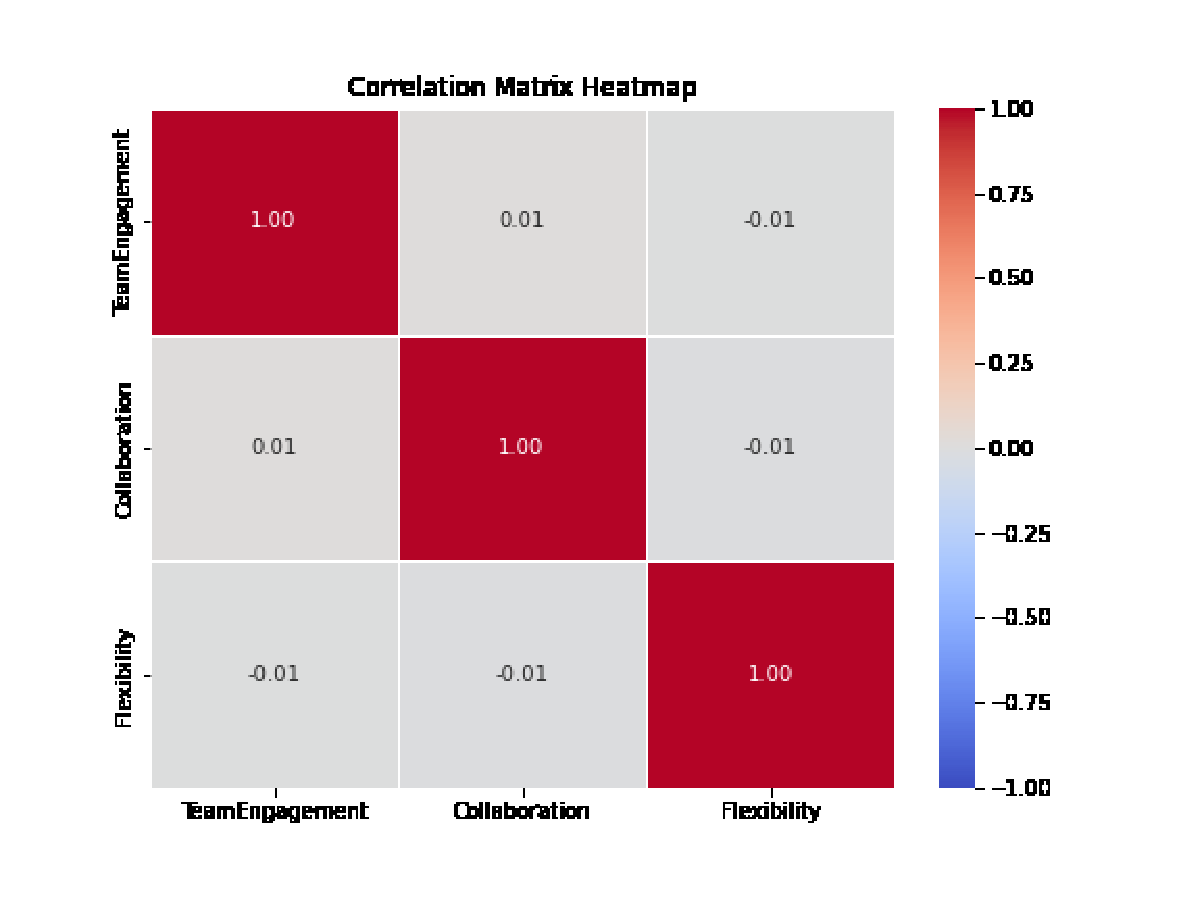}
  \caption{Heatmap for Bootstrapping}
\end{figure}

The heatmap shows that the synthetic data for Team Engagement, Collaboration, and Flexibility does not exhibit strong correlations. The correlations are close to zero, meaning these variables do not have significant linear relationships with one another in this dataset.

\subsubsection{Copula Method}
Copulas are functions that allow us to model the dependence between random variables separately from their marginal distribution. Employee behaviors like team engagement, collaboration, and flexibility are often interdependent, and copulas statistical methods for employee behavior are used. 
\[
U_i = \Phi(X_i) \quad \text{for each } i = 1, 2, 3
\]

\[
Y_i = \text{Beta}_{\alpha_i, \beta_i-1}(U_i) \quad \text{for each } i = 1, 2, 3
\]
\begin{figure}[H] 
  \centering
  \includegraphics[width=0.85\textwidth]{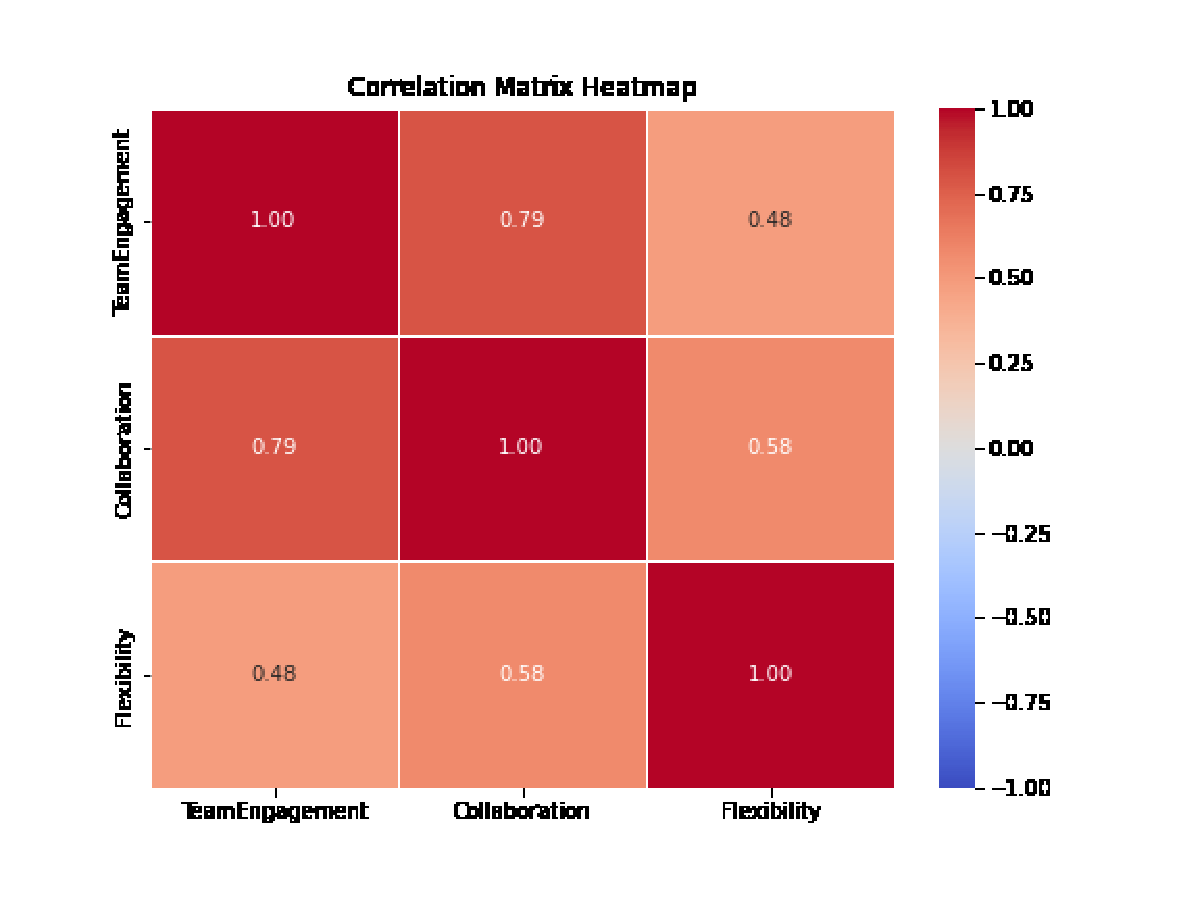}
  \caption{Heatmap for Copula}
\end{figure}
The heatmap shows strong positive correlations between employee behaviors, particularly between "Team Engagement" and "Collaboration" (0.79), indicating that engaged employees tend to collaborate more. The correlations between "Team Engagement" and "Flexibility" (0.48) and between "Collaboration" and "Flexibility" (0.58) are moderate, suggesting these behaviors are positively related but less strongly. This insight can guide targeted interventions to enhance engagement and collaboration while maintaining flexibility.

\subsection{Agent Based Method}
Agent-based modeling (ABM) is a simulation technique that can generate synthetic data for complex systems, such as organizational employee behavior. ABM involves creating agents (individual entities, such as employees) that interact according to a set of rules within a simulated environment.
\[
\text{Normalization: } \frac{P_i}{100}
\]
\[
\text{Behavior Metric calculation: } M_i = \min\left(\max\left(\mathcal{N}\left(\frac{P_i}{100}, 0.1\right), 0\right), 1\right) \times 100
\]
This approach allows for the generation of realistic synthetic data that reflects variations in employee behavior based on performance scores, which can be further analyzed or used for machine learning model training. The use of normally distributed random variables introduces a natural variability that mimics real-world differences in behavior.

\begin{figure}[H] 
  \centering
  \includegraphics[width=1.05\textwidth]{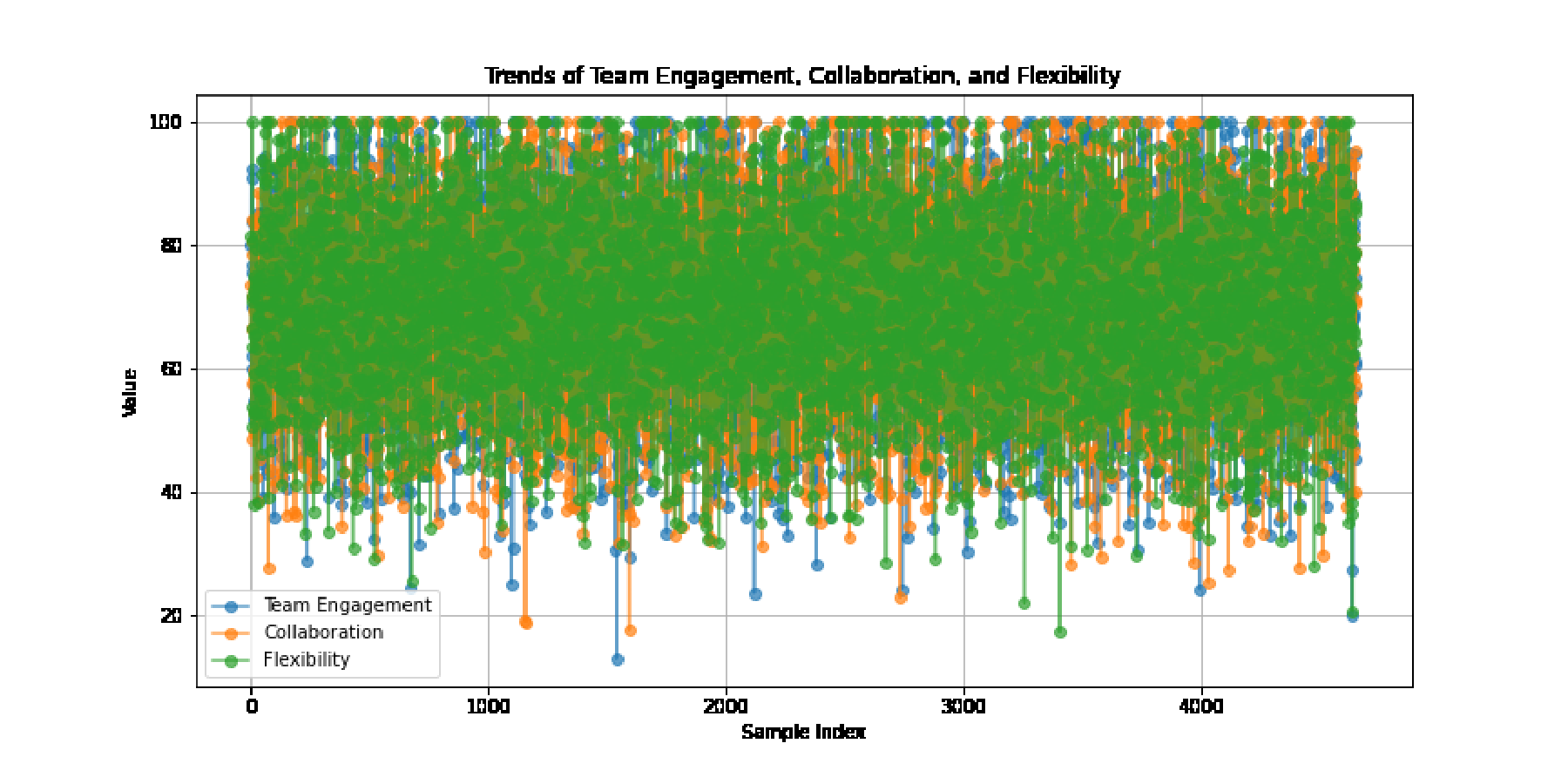}
  \caption{Trends Graph for Agent-Based Model}
\end{figure}

The graph shows the trends of three synthetic behavior metrics—Team Engagement, Collaboration, and Flexibility—generated using agent-based modeling (ABM).
The graph displays the distribution of synthetic behavior metrics—Team Engagement, Collaboration, and Flexibility—generated using agent-based modeling. Each metric is spread widely across the 0 to 100 range, with dense overlapping points indicating significant variability. The graph shows no clear trends, reflecting the random nature of the data generation process where each employee's behavior is independently influenced by their performance score. The overall pattern suggests a realistic simulation of diverse employee behaviors.

\subsection{Generative Adversarial Network}
Ian Goodfellow introduced it as a type of deep learning model used primarily for generating new data that mimics a given dataset. Through an adversarial process between a generator and a discriminator, GANs can produce realistic data often indistinguishable from real data. The training process of a GAN is adversarial because the generator and discriminator compete:
\begin{itemize}
    \item The generator tries to fool the discriminator by producing increasingly realistic data.
    \item The discriminator tries to improve its ability to detect fake data.
\end{itemize}
This process continues until the generator produces data indistinguishable from the real data.
Using a Generative Adversarial Network (GAN) to generate synthetic data for employee behavior offers several advantages, particularly in scenarios where data privacy, data scarcity, or the need for diversity and realism in data are key factors.
Generator(G)

\[Generator
G(z) = \sigma\left(W_3 \cdot \text{ReLU}\left(W_2 \cdot \text{ReLU}\left(W_1 \cdot z + b_1\right) + b_2\right) + b_3\right)
\]
Where z is the input noise vector, W1, W2, and W3 are the weight matrices,b1,b2, and b3 are the bias vectors, and ReLU(x)=max(0,x) is the activation function.

\[
D(x) = \sigma\left(W'_3 \cdot \text{ReLU}\left(W'_2 \cdot \text{ReLU}\left(W'_1 \cdot x + b'_1\right) + b'_2\right) + b'_3\right)
\]
Where x is data, W'1, W'2, W'3 are the weight matrices for the discriminator, B'1, B'2, B'3
\[
Discriminator Loss  = -\left[\log\left(D(x_{\text{real}})\right) + \log\left(1 - D(G(z))\right)\right]
\]
G(z) is the synthetic data generated by the generator, and D(x) is the discriminator's prediction.
\[
Generator Loss = -\log\left(D(G(z))\right)
\]
\[
\rho_{X,Y} = \frac{\text{Cov}(X,Y)}{\sigma_X \sigma_Y}
\]

\begin{figure}[H] 
  \centering
  \includegraphics[width=0.85\textwidth]{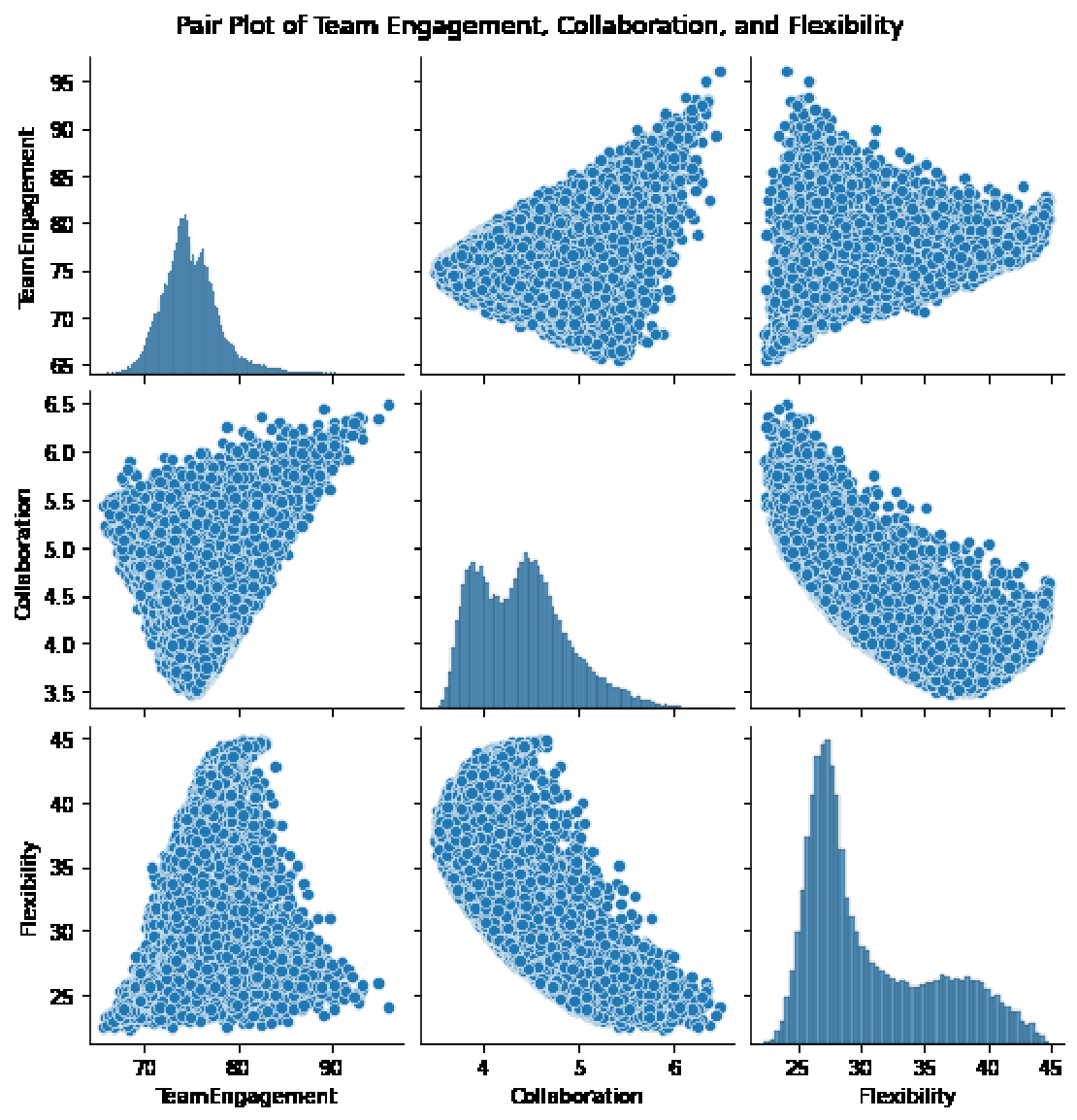}
  \caption{Pair plot for GAN}
\end{figure}
The pair plot demonstrates that the GAN effectively captured the distributions and complex, non-linear relationships between Team Engagement, Collaboration, and Flexibility. The histograms show that the synthetic data closely matches the original data's distribution. The scatter plots reveal realistic, diverse outputs, avoiding issues like mode collapse. Overall, the GAN-generated data maintains the integrity and variability of the original dataset, making it suitable for further analysis and data augmentation. This underscores the GAN's success in replicating key features of employee behavior metrics.

\section{Conclusion}
This study explores the creation and application of synthetic data for analyzing employee behavior, highlighting the significant role it plays in modern organizational management. By leveraging advanced techniques such as Generative Adversarial Networks (GANs), Agent-Based Models (ABMs), and statistical methods like copulas and bootstrapping, synthetic data can closely mimic real-world employee behavior while maintaining privacy and addressing data scarcity. The analysis of synthetic data offers valuable insights into key behavioral metrics such as performance, team engagement, collaboration, and flexibility, which are essential for enhancing productivity and workflow within organizations. The results demonstrate that synthetic data not only preserves the statistical characteristics of real data but also enables robust scenario testing and model training, making it an indispensable tool for employee behavior analysis and organizational efficiency improvement.

Here is a link to my GitHub repository: \href{https://github.com/rakshithajayashankar/SyntheticDataforEmployeeBehavior}{GitHub Repository}


\begin{thebibliography}{9}

\bibitem{jayashankar2024}
Jayashankar, R., \& Balan, M. (2024). Unveiling the dynamics of employee behavior through Wolfram's cellular automata. \textit{arXiv}. Retrieved from \url{https://arxiv.org/abs/2407.09581}

\bibitem{arimie}
Arimie, J. C., \& Oronsaye, A. O. Assessing employee relations and organizational performance: A literature review.
\bibitem{moez2024}
Moez, M. (2024). Synthetic Data is the Future of Artificial Intelligence. \textit{Medium}. Retrieved from \url{https://moez-62905.medium.com/synthetic-data-is-the-future-of-artificial-intelligence-6fcfd2ce1a14}

\bibitem{neuhausen2020}
Neuhausen, M., Herbers, P., \& König, M. (2020). Using Synthetic Data to Improve and Evaluate the Tracking Performance of Construction Workers on Site. \textit{Applied Sciences}, 10(14), 4948. Retrieved from \url{https://doi.org/10.3390/app10144948}

\bibitem{sdv2018}
\textit{SDV: Synthetic Data Vault}. (2018). Retrieved from \url{https://dai.lids.mit.edu/wp-content/uploads/2018/03/SDV.pdf}

\bibitem{howe2024}
Howe, B., Stoyanovich, J., Ping, H., Herman, B., \& Gee, M. \textit{Synthetic Data for Social Good}. University of Washington and Drexel University.




\bibitem{indikaai}
Synthetic Data Generation. (2024). \textit{IndikaAI Blog}. Retrieved from \url{https://www.indikaai.com/blog/synthetic-data-generation}



\end{thebibliography}
\end{document}